\documentclass{article}

\usepackage[preprint,nonatbib]{nips_2018}
\usepackage[utf8]{inputenc} % allow utf-8 input
\usepackage[T1]{fontenc}    % use 8-bit T1 fonts
\usepackage{hyperref}       % hyperlinks
\usepackage{url}            % simple URL typesetting
\usepackage{booktabs}       % professional-quality tables
\usepackage{amsfonts}       % blackboard math symbols
\usepackage{nicefrac}       % compact symbols for 1/2, etc.
\usepackage{microtype}      % microtypography

\usepackage[ruled]{algorithm2e}
\usepackage{algpseudocode}
\usepackage{multirow}
\usepackage{makecell}
\usepackage{amsmath}
\usepackage{pstricks}    %for embedding pspicture.
\usepackage{graphicx}    %for figure environment.
\usepackage{bbm}
\usepackage{enumerate}
\usepackage{subfigure}
\usepackage{url}

\title{Efficient learning of neighbor representations for boundary trees and forests}

\author{
  Tharindu Adikari \\
  University of Toronto \\
  \texttt{tharindu.adikari@mail.utoronto.ca} \\
   \And
   Stark C. Draper \\
   University of Toronto \\
   \texttt{stark.draper@utoronto.ca} \\
}

\begin{document}

\maketitle

\begin{abstract}
We introduce a semiparametric approach to neighbor-based classification. We build off the recently proposed \emph{Boundary Trees} algorithm by Mathy et al.~(2015) which enables fast neighbor-based classification, regression and retrieval in large datasets. While boundary trees use an Euclidean measure of similarity, the \emph{Differentiable Boundary Tree} algorithm by Zoran et al.~(2017) was introduced to learn low-dimensional representations of complex input data, on which semantic similarity can be calculated to train boundary trees. As is pointed out by its authors, the differentiable boundary tree approach contains a few limitations that prevents it from scaling to large datasets. In this paper, we introduce \emph{Differentiable Boundary Sets}, an algorithm that overcomes the computational issues of the differentiable boundary tree scheme and also improves its classification accuracy and data representability. Our algorithm is efficiently implementable with existing tools and offers a significant reduction in training time. We test and compare the algorithms on the well known MNIST handwritten digits dataset and the newer Fashion-MNIST dataset by Xiao et al.~(2017).
\end{abstract}

\section{Introduction}
%%%%%%%%%%%%%%%%%%%%%%%%%%%%%%%%%%%%%%%%%%%%%%%%%%%%%%%%%%%%%%%%%%%%%%%%%%%%%%%%%

%Recent advances in machine learning have drastically changed our way of life. The ability of machine learning algorithms and tools to handle large volumes of data have helped us build complex systems that control almost all aspects of our everyday life.

Neighbor-based classification methods have long been used and have been well studied. They have received somewhat less attention in the past few years due to advances in rival algorithms such as support vector machines and neural networks. However, neighbor-based methods are making a comeback as they have become feasible for many applications, due in no small part to fast neighbor search algorithms that scale to high-dimensional massive datasets (\cite{mathy2015boundary}, \cite{li2016fast}, \cite{andoni2015optimal}). Although neighbor-based methods are not always the best performing in terms of accuracy, they have commendable properties that other methods do not always posses. Neighbor-based methods are nonparametric and let data directly drive predictions without making strong modeling assumptions~\cite{chen2018explaining}. In addition to generating predictions, a neighbor-based classifier is able to provide similar instances to a given query point, which can be used to help explain its decision making process~\cite{lipton2016mythos}. This is akin to how humans sometimes justify decisions - by analogy. For example, it can help a medical practitioner to decide which treatment to prescribe to a patient if the practitioner is provided with examples of the raw data of patients that had similar symptoms. This can be a significant advantage of neighbor-based methods in certain contexts when compared to black-box type classifiers such as neural networks.

%Despite the many uses of the automated systems, lack of their interepretability has hindered them being adopted in critical areas like medicine, the criminal justice system, and financial markets.

% The literature suggest that there is no unified notion of what is meant by interepretability in the context of machine learning. Collectively, the literature refer to interpreting the learning model itself or the training method or the result.

%Our method leverages on insights provided in two recently proposed ideas which we present below. In this paper we introduce a scheme, that helps understand how that result was obtained and why. Our method attempts to justify its output by providing semantically similar training examples that led to a particular output. For example, a diagnosis model might provide intuition to a human decision-maker by pointing to similar cases in support of a diagnostic decision.

% Our work presented in this paper is done on top of 
In this paper, we present an approach to efficient approximate nearest neighbor methods inspired by the recently proposed \emph{Boundary Tree}~\cite{mathy2015boundary} and \emph{Differentiable Boundary Tree}~\cite{zoran2017learning} algorithms. The boundary tree (BT) algorithm is a nonparametric, computationally fast, and memory-efficient approach for neighbor-based classification, regression and retrieval problems. The BT algorithm is able to learn in an online fashion where training data is made available incrementally. Like any other neighbor-based method, the boundary tree algorithm requires a metric that measures semantic similarity of training data. The differentiable boundary tree (DBT) algorithm generalizes the boundary tree idea by, in addition, learning representations of training data in a supervised fashion, on which semantic similarity can be approximated. The DBT algorithm learns a transformation function in conjunction with a BT, therefore, can be considered `semiparametric'. The learned representations are then used to compute pairwise similarity of data, towards building a boundary tree. However, as noted by the original authors, batching is infeasible for the proposed DBT scheme. This limitation prevents it from exploiting fully the benefits offered by modern machine learning tools such as Graphics Processing Units (GPUs) and Tensor Processing Units (TPUs)~\cite{jouppi2017datacenter}, which are heavily reliant on batch-implementation. 
%In case a boundary tree is trained and some previously unseen data has been available, the BT can be re-trained on new data.

%We analyze and apply the useful ideas taken from the DBT method. 
In this paper, we present an algorithm that overcomes the computational issues of the DBT approach and also improves accuracy and data representability. Our algorithm is efficiently implementable with existing tools and offers a significant improvement in convergence time.
%We also extend the proposed algorithm to semi-supervised learning by making use of the recently proposed \emph{ladder networks}~\cite{rasmus2015semi}. In this setting, neighbor representations are learned using only a small number of labeled exemplars, along with a large number of unlabeled data.%, where a small set of labeled data is used to.

%++ need a third para to conclude into. what is done in this paper and why, results? can perhaps have a for outline

\newcommand{\Rel}{\mathbb{R}}
\newcommand{\vv}[1]{\mathbf{#1}}
\newcommand{\vx}[1]{\vv{x}_{#1}}
\newcommand{\distx}[2]{d(\vx{#1}, \vx{#2})}
\newcommand{\cc}[1]{\mathcal{#1}}
\newcommand{\setX}{\cc{X}}
\newcommand{\setY}{\cc{Y}}
\newcommand{\setD}{\cc{D}}

\section{Background}
%%%%%%%%%%%%%%%%%%%%%%%%%%%%%%%%%%%%%%%%%%%%%%%%%%%%%%%%%%%%%%%%%%%%%%%%%%%%%%%%%

%Motivation as an interpretable result. Neighbor based approach.
%an input $\vv{x}$ output mapping happens only through the.
Throughout this paper we use $[K]$ to denote the index set $\{1, \dots, K\}$ for any positive integer $K$. Let $\setD=\{(\vx{n}, y_n)\mid \vx{n} \in\Rel^D, y_n\in[C], n\in[N]\}$ be a dataset where $y_n$ is the label of $n$-th data point $\vx{n}$ and $C$ is the number of classes. We define $\setX=\{\vx{n} \mid n\in[N]\}$ and $\setY=\{y_n\mid n\in[N]\}$. Also, we use $\lVert \cdot \rVert$ to denote the $L_2$ norm.

\subsection{Nearest-neighbor classification}

Given a training dataset $\setD$ and query data point $\vx{}$, the simplest version of neighbor-based classification, known as $1$-nearest neighbor ($1$-NN), is to search the set $\setX$ for the closest point to $\vx{}$ and assign the label of the search result as the predicted label for $\vx{}$. %Concisely, the predicted label for $\vx{}$ is $y_{n^*}$ where %$n^*$ is the index of the vector in $\setX$ that is closest to $\vx{}$. 
%\[n^* = \argmin_{n \in [N]} \lVert \vx{} - \vx{n} \rVert.\]
A generalization of the $1$-NN method is $k$-NN, where instead of one, the $k$ closest data points of $\vx{}$ are selected and, from their associated labels, the most common label is taken as the prediction. One major drawback of the $k$-NN method is that it is not scalable to large and high dimensional datasets. To identify the closest $k$ data points of $\vx{}$, the Euclidean distance has to be computed from $\vx{}$ to each element of $\setX$ and the set of computed distances must be searched for the smallest $k$ among them. The distance calculations and search operations can be computationally infeasible when $D$ and $N$ are large.

\newcommand{\calT}{\cc{T}}

\subsection{Boundary trees and forests} \label{sec:btf}
The boundary tree (BT) algorithm described in~\cite{mathy2015boundary} elegantly overcomes the scalability issues of $k$-NN. When presented a large dataset, the BT algorithm learns a small subset of data points that is able to adequately represent the whole dataset. The proposed scheme is built around a tree structure $\calT$, consisting of nodes and edges. Each node of $\calT$ uniquely represents some data point/label pair. In order to perform classification, the tree is first trained considering $\setD$ in a sequential manner. In the beginning $\calT$ is empty and some random data point/label pair from $\setD$ is set as the root node of $\calT$. Since we have no prior knowledge about the order of the elements in $\setD$, without loss of generality, we assume $(\vx{1}, y_1)$ is set as the root. Given any subsequent training pair $(\vx{i}, y_i)$ $i\in\{2,\dots,N\}$, $\calT$ is traversed starting from the root, searching for the node with the closest data point to $\vx{i}$. At each step the data point of the current node and the data points of all its children are considered. If the closest node happens to be a child of the current node, the search is recursively repeated by setting the child to be the current node. Otherwise, the search ends and the current node is taken as the node in the tree that is (approximately) closest to $\vx{i}$. Assuming the tree traversal returns a node pair $(\vx{j}, y_j)$, then the training point $(\vx{i}, y_i)$ is discarded if $y_i=y_j$ as this training point is already sufficiently well represented by $\calT$. On the other hand, if $y_i\neq y_j$ the training point is added to the tree as a child node of $(\vx{j}, y_j)$.% Note that the dataset $\setD$ is considered sequentially, not all the data is required training happens in a  in the training phase, as data become available later.

After training, $\calT$ can be used for what is called approximate nearest neighbor (ANN) classification. ANN classification is different from $k$-NN. Whereas the latter considers the true nearest neighbors, the former does not always consider the true nearest neighbors but rather data points in the neighborhood (an \emph{approximation}) of the true nearest neighbors. In the BT algorithm, if we are given query point $\vx{}$, classification is performed by first searching $\calT$ for the (approximately) closest point to $\vx{}$. The same search scheme is used as during training. The data point returned by the search is selected as the ANN of $\vx{}$, and its label is assigned to $\vx{}$ as the classification result. By construction, the boundary tree has the property that any two connected nodes belong to different classes. Therefore, the majority of the nodes in the tree will be near class boundaries. One can think of the BT algorithm as a combination of two concepts. The first is the selection of data points near class boundaries. The second is to make use of a tree structure that enables efficient search for the ANN of a given data point. The authors of the BT algorithm also propose a mechanism to restrict the number of children at any node. This controls the expansion of the tree at any given depth. The query and training time of the resulting tree scale as $\log(N)$ and $N\log(N)$ respectively.

A boundary forest (BF) is an ensemble of BTs trained on different permutations of a given dataset. Classification with a BF is performed by combining the predicted label of each tree according to some weighting scheme. In our work we focus on BTs, but our results are applicable to BFs as well.

%The nearest-neighbor of $\vx{}$, returned by BT, might not be the closest data point in $\setX$ to $\vx{}$. This is partly because, the tree does not contain all the data points provided in $\setX$, and there is a chance that the closest data point to $\vx{}$ in $\setX$ is left out from the tree. Another reason is, even if the tree contains the closest data point of $\vx{}$, the search algorithm might not -- % As pointed out earlier, BT will only return an approximate neighbor of the query data point. However, the BT algorithm has the property, if $m$ is the smallest fraction such that the returned approximate nearest neighbor is included in true $mN$-nearest neighbors, $f$ decreases in $N$. % if training dataset is large, with high probability, we refer to the original paper for details.

%Boundary tree allow for fast k-NN classification, regression and retrieval and their memory requirements grow very slowly with the amount of data presented, all within a simple and elegant formulation.

% This algorithm has its varient is tweaked In this paper we only deal with the case of classification although regression and retrieval are also possible.

%\cite{mathy2015boundary}
%Mathy et al. 2015 proposed the Boundary Forest Algorithm that summarizes a large dataset using a small number of examples.

%\newcommand{\bchT}{\cc{B}_{\calT}}
%\newcommand{\bchf}{\cc{B}_{f}}
\newcommand{\setU}{\cc{U}}
\newcommand{\fnt}[1]{f_{\theta}(#1)}
\newcommand{\fntd}{\fnt{\cdot}}

\newcommand{\sztrn}{N_{\text{t}}}
\newcommand{\szbnd}{N_{\text{b}}}

\subsection{Differentiable boundary trees} \label{sec:dbt}
In the BT algorithm described above, Euclidean distance was assumed. That is, given two data points $\vx{i}$ and $\vx{j}$, the similarity between them equals the distance $\distx{i}{j} = \lVert \vx{i} - \vx{j} \rVert$. While acceptable in certain datasets, Euclidean distance may not be a good choice for complex data such as images. As is proposed in~\cite{mathy2015boundary}, this limitation can be mitigated by first transforming the input data using a known feature-extraction function $f(\cdot)$ and, subsequently, measuring Euclidean distance between extracted features. In this case, the distance between $\vx{i}$ and $\vx{j}$ is $\distx{i}{j} = \lVert f(\vx{i}) - f(\vx{j}) \rVert$. One drawback of this proposal is that $f(\cdot)$ needs to be carefully crafted, crafting that might require expert, domain-specific knowledge. In~\cite{zoran2017learning}, the authors aim to assuage this limitation of boundary trees. They describe an algorithm called the differentiable boundary tree (DBT) algorithm, which essentially \emph{learns} a transformation function $\fntd$, parameterized by $\theta$, such that Euclidean distance is a good distance metric in the transformed domain. The function $\fntd$ is implemented by a neural network (NNet) and the dimension of the output (the range of $\fntd$) is kept small relative to that of the input. The parameter set $\theta$ is initialized randomly and is jointly optimized with a boundary tree using stochastic gradient descent (SGD) as per the scheme we describe next.
%Optimizing $\theta$ and training the boundary tree happen in an iterative manner which we summarize below.

\newcommand{\setDb}{\bar{\cc{D}}}
\newcommand{\setUb}{\cc{U}_{\text{b}}}
\newcommand{\setV}{\cc{V}}
\newcommand{\indctr}{\mathbbm{1}}

\newcommand{\node}[1]{\text{node-}{#1}}
\newcommand{\parent}[1]{\text{p}({#1})}
\newcommand{\sibls}[1]{\mathcal{W}({#1})}
\newcommand{\indq}{r}
\newcommand{\indlast}{s}

Given training dataset $\setD$, let $\setDb$ be a subset (a \emph{mini-batch}) of size $\szbnd+1$ $(\ll N)$ and $\setU = \{(\fnt{\vx{i}}, y_i) \mid (\vx{i}, y_i)\in \setDb\}$. First, we build a boundary tree $\calT$, using the first $\szbnd$ elements of $\setU$. We use $\node{i}$ to indicate the node in $\calT$ that consists of the pair $(\fnt{\vx{i}}, y_i)$. Also, let $\parent{i}$ be the parent node index of $\node{i}$, and $\sibls{i}$ be the index set of its siblings. In the second step, we use the last element of $\setU$, i.e., the $\szbnd+1$st element, to query $\calT$. We assume the index of this element into $\setD$ is $\indq\in[N]$. The boundary tree $\calT$ is traversed to get the ANN of $\fnt{\vx{\indq}}$, as described in Section~\ref{sec:btf}. Note that the tree traversal is a series of transitions, each a transition from a parent node to a child. A set $\setV$ is maintained to keep track of the tree traversal. The node index $i$ is included in $\setV$ if the traversal includes a transition to $\node{i}$ from its parent. Observing that each transition is conditionally independent of previous transitions, the authors of DBT assert a probabilistic model on the path traversed. As per the model, recalling that $\vx{\indq}$ is the query point and $\vx{j}$ is the data point associated with node-$j$, the probability of transitioning to a $\node{i}$ from its parent is
\begin{equation}\label{eq:prtrans}
\Pr(\parent{i}\to i | \indq) = \frac{\exp(-\distx{i}{\indq})}{\sum_{j\in\{i,\parent{i}\}\cup\sibls{i}}\exp(-\distx{j}{\indq})},
\end{equation}
which results in the log probability of traversing the observed path 
$\log\Pr(\text{observed-path}) = \sum_{i\in\setV}\log \Pr(\parent{i}\to i | \indq).$ 
In our experiments, presented in Section~\ref{sec:expr}, we replace $d(\cdot, \cdot)$ in (\ref{eq:prtrans}) by $d(\cdot, \cdot)/\sigma$, allowing the positive scalar value $\sigma$ to control the sensitivity of the distance measure. A soft class prediction on $\vx{\indq}$ is obtained by \emph{separately} considering the final transition. By letting $\node{\indlast}$ be the last node in the tree traversal, the (unnormalized) log probability of $\vx{\indq}$ belonging to class $c\in[C]$ is % which is returned as the ANN of $\vx{\indq}$,
\begin{equation}\label{eq:pryc}
\log{\Pr}^*(y_{\indq}=c) = \Bigg[\sum_{i\in\setV\setminus \indlast}   \log\Pr(\parent{i}\to i | \indq) \Bigg]     +
 \log  \Bigg[ \sum_{i\in\sibls{\indlast}\cup\{\indlast\}} \Pr(\parent{i}\to i | \indq)\indctr[y_i=c]\Bigg],
\end{equation}
where $\indctr[\cdot]$ denotes the indicator function. The computed class probabilities are normalized to obtain a proper distribution, which produces $\Pr(y_{\indq}=c) = {\Pr}^*(y_{\indq}=c)\big/\sum_{c'\in[C]}{\Pr}^*(y_{\indq}=c')$. Though the authors of~\cite{zoran2017learning} have not noted the following, we observe that the contribution from the first term in (\ref{eq:pryc}) completely disappears after normalizing, nullifying the effect of this term on the rest of the computations. The significance of this observation will become clear when we motivate our own algorithm development in Section~\ref{sec:motivation}. After obtaining soft class predictions $\Pr(y_{\indq}=c)$, the cross entropy loss $\cc{L}$ is calculated using the true label $y_{\indq}$ and $\Pr(y_{\indq}=c)$. Finally, the gradient of loss $\nabla_{\theta}\cc{L}$ is computed and one gradient step is taken towards minimizing $\cc{L}$. The process of building $\calT$ and minimizing $\cc{L}$ is repeated till the loss converges, using a new mini-batch $\setDb$ each time. The convergence rate can be increased by considering a mini-batch of size $\szbnd+\sztrn$ $(\ll N)$ instead of $\szbnd+1$. In each iteration, the first $\szbnd$ elements are used to build $\calT$. The rest of the elements are then used to take $\sztrn$ sequential gradient steps towards minimizing $\cc{L}$.

%taking one gradient step using one $(\fnt{\vx{l}}, y_{\indq})$ sample in each iteration, $\sztrn$ number of gradient steps are taken with $\sztrn$ number of such samples.

\subsection{Related work}
Nearest-neighbor (NN) methods have been around for a long time and many different variants have been studied. As noted in~\cite{zoran2017learning}, these methods can be broadly categorized into tree-based methods and hashing-based methods. The former leverage tree structures to search NNs efficiently, drastically reducing search time in the massive datasets typical of contemporary applications. Hashing-based methods like those presented in~\cite{andoni2015optimal} and~\cite{salakhutdinov2009semantic} better suit high-dimensional data. This family of methods typically compute a low-dimensional hash value for given data points and search for NNs based on the proximity of hashes. Although the DBT scheme and the algorithm we present next are mostly related to the tree-based methods, we note that they learn functions that transform high-dimensional data to a low-dimensional space, and thus make a connection to hashing-based methods as well.

The work done in~\cite{koch2015siamese} and~\cite{hoffer2015deep} are the closest to the DBT algorithm in terms of learning a representation of the data. Siamese networks are proposed in~\cite{koch2015siamese} to classify any given pair of data points as coming from the same or from different classes. In the simplest variant of the Siamese networks, given $\vx{i}$ and $\vx{j}$, a function $\fntd$ is learned ensuring that $\lVert \fnt{\vx{i}} - \fnt{\vx{j}} \rVert$ is small if $y_i=y_j$, and large otherwise. The triplet networks proposed in~\cite{hoffer2015deep} operate similarly, but use a slightly different training scheme. Given a triplet $(\vx{i}, \vx{j}, \vx{l})$ with the property that $y_i=y_j\neq y_l$, a triplet network is trained to learn a function $\fntd$ such that the inequality $\lVert \fnt{\vx{i}} - \fnt{\vx{j}} \rVert \ll \lVert \fnt{\vx{i}} - \fnt{\vx{l}} \rVert$ is satisfied. We refer the reader to~\cite{zoran2017learning} for a brief summary on other DBT-related work. What makes the DBT algorithm unique among existing representation learning methods is that it learns a function that is specifically trained to build BTs. In contrast, one could propose to learn a transformation $\fntd$ with a Siamese or triplet network, and build a BT using the learned transformation. But if this strategy is followed, it means that $\fntd$ does not learn to fully appreciate the importance of points near class boundaries, points that are the most important in building BTs. In Figure~\ref{fig:bnd_impct}, we illustrate the importance of using points near class boundaries in learning a function $\fntd$. We further discuss the significance of this concern in Section~\ref{sec:motivation}.

\newcommand{\clsg}{\texttt{green}}
\newcommand{\clsr}{\texttt{red}}

\begin{figure}
	\centering{
		\includegraphics[width=0.35\textwidth]{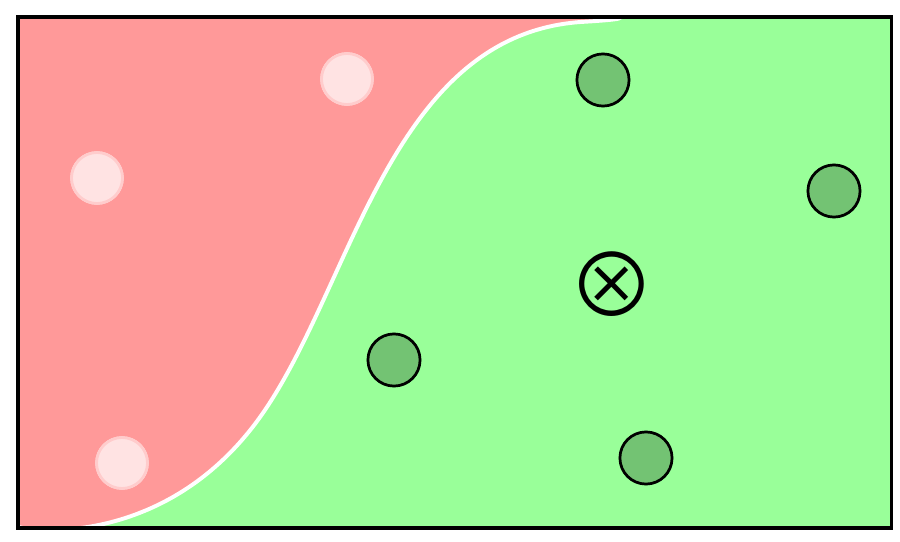}
		\includegraphics[width=0.35\textwidth]{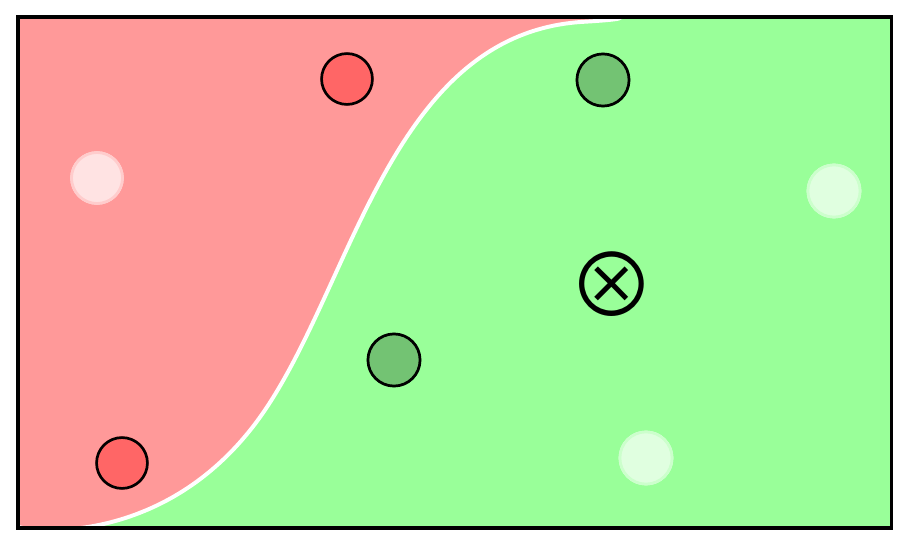}
		%\resizebox{0.5\textwidth}{!}{\input{artwork/bnd_bg}}
	}
	%\fbox{\rule[-.5cm]{0cm}{4cm} \rule[-.5cm]{4cm}{0cm}}
	\caption{Illustrating the importance of selecting points near class boundaries when learning $\fntd$. 
		Each sub-figure portrays the same underlying problem with a boundary between two classes $\clsr$ and $\clsg$. A few data points, marked by empty circles, lie on the correct sides of the boundary. In this example, we will ignore the tree structure for clarity, but the conclusions pertain to BTs. Given the query data point $\vx{}$, marked by the circled cross, we consider its 4-NNs (with no regard to the boundary) as illustrated in the left-side figure. In this case, the class prediction already has a high confidence that $\vx{}$ belongs to $\clsg$ class since it is closer to and is surrounded by points in said class. Such a prediction offers little incentive to separate further the two classes. The figure in right-side demonstrates the case when the 4-NNs that are also \emph{near the boundary} are considered. In this case, class prediction has a low confidence that $\vx{}$ belongs to $\clsg$ class, which incentivizes $\fntd$ to learn to separate the two classes and to pull the $\clsg$ points away from the boundary.}\label{fig:bnd_impct}
\end{figure}

\section{Algorithm}
%%%%%%%%%%%%%%%%%%%%%%%%%%%%%%%%%%%%%%%%%%%%%%%%%%%%%%%%%%%%%%%%%%%%%%%%%%%%%%%%%

The key contributions provided by the BT and DBT algorithms can be summarized as follows. The BT algorithm offers fast search for approximate nearest neighbors among a large number of high dimensional data points. Applications include classification, regression and data retrieval. The DBT algorithm learns a function $\fntd$ that transforms high dimensional data to a space where Euclidean distance is a good measure of similarity with respect to labels. In this section we first make a number of observations about BTs and the DBT scheme, which motivate our new algorithms \emph{Boundary Sets} and \emph{Differentiable Boundary Sets}. We then describe the proposed algorithms.% in succeeding subsections.

\subsection{Motivation} \label{sec:motivation}

With regard to BT and DBT algorithms we make following observations and deductions that motivate us to alter the DBT scheme.

\begin{enumerate}[(i)]

\item
As mentioned in Section~\ref{sec:dbt}, the contribution from the first term in (\ref{eq:pryc}) disappears after normalization and the term has no effect on the loss $\cc{L}$. This suggests that only a small set of siblings of the ANN of a given training data point actually contribute to learning $\fntd$. Most of the data points in the tree, which otherwise could have helped the learning process, are completely ignored.

\item \label{it:upperbnd}
The number of nodes in $\calT$ is upper bounded by $\szbnd$, the number of examples used to train the tree. In the experiments performed in~\cite{zoran2017learning} on MNIST and CIFAR10 datasets, $\szbnd$ is set to 1000 and 100 respectively, which are relatively small compared to the dataset sizes.%, which are relatively .% Evidently, over the time, the size of the tree becomes smaller and smaller ~22 even for a complex dataset like CIFAR10. Since we're using MB anyway, it makes sense to get rid of the tree in the training phase.

\item \label{it:simpletrans}
%In the transformed domain, the dimension of data can be relatively small .
The transformation $\fntd$ throws away a lot of information that is unrelated to labels and simplifies class boundaries. As a direct result, often only a small number of data points need to be included in the tree. This can be observed in the experiments performed in~\cite{zoran2017learning} where the boundary tree learned for MNIST dataset is represented by only 25 samples and that for the CIFAR10 dataset by only 22 samples.

\item
Observations (\ref{it:upperbnd}) and (\ref{it:simpletrans}) indicate that, while training the DBT, the size of the tree is within our control, and often falls below 30 even for real world datasets. Since the true benefit of the tree structure, which is efficient search, emerges only when the stored number of samples is large, then there is little reason to believe that such tree structure has a significant impact on the DBT algorithm. %dataset is large, it might not have a little .

\item \label{it:anyfuncokay}
The original motivation to develop the DBT algorithm is to learn a feature-extraction function $\fntd$ such that $\distx{i}{j} = \lVert \fnt{\vx{i}} - \fnt{\vx{j}} \rVert$ is a good measure of similarity between the labels of $\vx{i}$ and $\vx{j}$. Subsequently, the learned function can be used to transform the given dataset into the new feature space to build a BT or BF for the intended application. To this end, any method that efficiently learns an $\fntd$ with the desirable properties is applicable. There is no requirement that the training should involve a tree structure.
%boundary tree algorithm will work as long as distance between data points are measured after any transformation  %is  f() is a transformation such that dij = fx -fx is a good similarity measure.

\item
The observation made in (\ref{it:anyfuncokay}) provides an inkling that one may not need to use a tree structure in the DBT algorithm, rather using data points \emph{near class boundaries} is what has a significant impact on learning $\fntd$. This is because points near a class boundary are the furthest from the cluster center, and closest to the boundary of another class. Given a random query data point $\vx{}$, its label $y$ can be predicted by either considering the ANNs of $\vx{}$, or considering the ANNs of $\vx{}$ that are \emph{also} near the class boundary. With regard to the latter, the predicted label is most likely to be weak or erroneous. Consequently, $\fntd$ is forced to make such weak predictions stronger by learning clusters whose class boundaries are well separated. This can be thought of as learning in an adversarial setting, where the provided neighbor points are carefully selected by an adversary to be the poorest choice among the ANNs. Thus making the learned classifier more robust as we illustrate in Figure~\ref{fig:bnd_impct}.

% the ANNs provided to make predictions are the most  worst choice. the points near boundary to cluster, and more importantly move away from points belonging to other classes. Using data points closer to the class boundaries instead of ones well within the boundaries for nearest neighbor-based classification. 

%If tree is let go, why boundary points? why not some random set of points? ans: ++ Describe why use examples on the boundary? Why not some random 1000 points? Why not use all 1000 (or 100) samples to predict label for query point? Why select a boundary set which is a subset of above 1000? This way, samples in the boundary are selected. So minimizing loss will encourage the training point to move away from the boundary towards the center of the cluster. Also, if the boundary samples are also function of theta, the boundary also will try to move away from other classes. Illustrate this with a figure.
%++ Why is above second case the model is performing better than classic NN?

\item
State-of-the-art neural network learning tools such as GPUs and TPUs~\cite{jouppi2017datacenter} use batch-wise optimization to scale to large datasets. As pointed out in~\cite{zoran2017learning}, many practical issues arise when a function $\fntd$ is optimized in conjunction with a tree, mainly because the depth of tree traversal and the number of ANNs differ from one training point to another. This causes the batch-implementation of DBT to be inefficient and prevents the algorithm from leveraging the benefits offered by modern tools that work best with batch-implementation.

%It might not be efficient to use a tree structure to represent such a small number of samples. due to limitations of the tools. 
\end{enumerate}

Based on the above observations, we argue that the difficulties due to the joint optimization of $\fntd$ with a tree structure outweigh the benefits. We stress the necessity to revamp the DBT algorithm or to look for an alternative, that not only serves its original purpose but also is efficient and conforms to modern machine learning techniques. To overcome the difficulties inherent to the DBTs, we propose the boundary \emph{set} and differentiable boundary \emph{set} algorithms which we present next.%fits well to growing demands of modern applications.

\newcommand{\calS}{\cc{S}}

\subsection{Boundary sets}

The boundary set (BS) is, in effect, a modified version of a boundary tree, including modifications geared toward efficiently learning a transformation function $\fntd$. The pivotal change in a boundary set is that it accumulates samples in a set $\calS$ without building a tree structure. Given the training dataset $\setD$, the BS considers data points sequentially, similar to BT. In the beginning $\calS$ is empty and $(\vx{1}, y_1)$ is added as its first element. Given any subsequent training pair $(\vx{i}, y_i)$, $i\in\{2,\dots,N\}$, elements of $\calS$ are searched to obtain the pair closest to $\vx{i}$, according to the distance function $\distx{i}{j}$. Assuming the search returns pair $(\vx{j}, y_j)$, the training point $(\vx{i}, y_i)$ is discarded if $y_i=y_j$, otherwise $(\vx{i}, y_i)$ is added to $\calS$. Clearly for the boundary sets to be useful the final set cannot be of too large a cardinality, else we would suffer from the same effects that first motivated the development of boundary trees.

\subsection{Differentiable boundary set algorithm} \label{sec:dbs}

We propose the differentiable boundary set (DBS) algorithm which addresses issues of DBT and improves its accuracy, convergence rate and data representability. Furthermore, DBS is easily and efficiently implementable with currently available software packages, and scales well to large datasets. Similar to DBT, the end objective of DBS is to learn a function $\fntd$. The difference in DBS is that, instead of a boundary tree, $\fntd$ is jointly optimized with a boundary set. With the new approach, the label prediction for a given query point $\vx{}$ is computed with respect to all elements in $\calS$. In contrast, DBT only considers data points that are in the tree traversal path (as was described in Section~\ref{sec:dbt}). This modification allows efficient implementation of the algorithm and also increases prediction accuracy. Given a training dataset $\setD$, a mini-batch size $\szbnd$ and a positive scalar value $\sigma$, Algorithm~\ref{algo:proposed} outlines the DBS training procedure.

%The DBS training scheme is similar to that of DBT, except

%\newcommand{\mbBt}{\cc{B}_{f}}
%\newcommand{\Yf}{Y_{f}}
%\newcommand{\Yhf}{\hat{Y}_{f}}
\newcommand{\Yb}{Y}%_{\text{b}}

%We argue that once the dimension of data is reduced, Our algorithm centers 
\begin{algorithm}[H]
	\caption{DBS training algorithm}\label{algo:proposed}
	%\KwData{this text}
	%\KwResult{how to write algorithm with \LaTeX2e }
	\SetKwInOut{Input}{input}
	\Input{$\setD, \szbnd, \sigma$}
	randomly initialize $\theta$, parameters of $\fntd$\;
	
	%https://tex.stackexchange.com/questions/142922/how-to-align-text-within-an-algorithm-environment
	\While{not reached maximum number of epochs}{
		shuffle elements and partition $\setD$ to obtain subsets of size $(\szbnd+1)$\;
		%\tcc{iterate over all training examples}
		\ForEach{subset $\setDb$}{
			$\setU \gets \{(\fnt{\vx{n}}, y_n) \mid (\vx{n}, y_n)\in \setDb\}$\;
			$\setUb \gets$ first $\szbnd$ elements of $\setU$\;
			$\calS \gets$ boundary set computed using elements of $\setUb$\;
			
			$(\fnt{\vx{\indq}},y_{\indq}) \gets$ last element of $\setU$\;
			$\vv{d} \gets$ row vector consisting Euclidean distances between each data point in $\calS$ and $\vx{\indq}$\;
			
			$\vv{h} \gets$ softmax function applied on $\frac{-\vv{d}}{\sigma}$ i.e., $\vv{h}(i) = \frac{\exp(-\vv{d}(i)/\sigma)}{\sum\limits_{j\in[|\calS|]}\exp(-\vv{d}(j)/\sigma)}$ for $i\in[|\calS|]$\;
			
			$\Yb \gets$ $|\calS| \times C$ matrix where rows are the one-hot label encodings of elements of $\calS$\;
			
			$\hat{\vv{y}} \gets \vv{h}\Yb$ where $\hat{\vv{y}}(c) = \Pr(y_{\indq}=c|\vx{\indq},\calS,\theta)$ for $c\in[C]$\;
			$\cc{L} \gets$ cross entropy loss calculated with $\hat{\vv{y}}$ and $y_{\indq}$\;
			
			Compute $\nabla_{\theta}\cc{L}$ and take one step to minimize $\cc{L}$\;
		}
	}
\end{algorithm}

%the vector $\vv{h}$ represents how $\vx{\indq}$ and elements of $\calS$ are related. Specifically, 
In Algorithm~\ref{algo:proposed}, the $i$-th component in $\vv{h}$ is a measure of the \emph{closeness} of $\vx{\indq}$ to the $i$-th data point in $\calS$, where the sensitivity of the measure is controlled by $\sigma$. The vector-matrix product $\vv{h}\Yb$ compactly computes the label prediction for $\vx{\indq}$, by summing the one-hot label encodings of elements in $\calS$, weighted by their closeness to $\vx{\indq}$. The vector $\hat{\vv{y}}$ is a vector of the probabilities that $\vx{\indq}$ belongs to each class. The algorithm can be efficiently implemented with batch support where the $(\fnt{\vx{\indq}},y_{\indq})$ pair is replaced with a size $\sztrn$ mini-batch. The Euclidean distance calculation step remains the most costly operation of the algorithm. However its computational complexity is now within our control as the size of the set $\calS$ is upper bounded by $\szbnd$. After learning $\fntd$, a \emph{final} boundary tree or a boundary forest is built with \emph{all} available training data using $\distx{i}{j} = \lVert \fnt{\vx{i}} - \fnt{\vx{j}} \rVert$ as the distance measure.

\section{Experimental results} \label{sec:expr}
%%%%%%%%%%%%%%%%%%%%%%%%%%%%%%%%%%%%%%%%%%%%%%%%%%%%%%%%%%%%%%%%%%%%%%%%%%%%%%%%%

In this section we test and compare the performance of the DBS and DBT algorithms. We use the well-known MNIST dataset of handwritten digits (`Digit-MNIST') and the newer `Fashion-MNIST'~\cite{xiao2017fashion} dataset. The images in Fashion-MNIST dataset are of fashion material such as cloths bags and boots. It is similar to Digit-MNIST in terms of image size, number of classes, and amount of training and test data, but classification is more challenging. %for Fashion-MNIST than it is for Digit-MNIST. 
Each dataset consists of 60000 training images that are of size 28$\times$28 and belong to 10 classes. Both datasets have separate test sets of 10000 images.

As per the DBT training procedure described in Section~\ref{sec:dbt}, given a mini-batch of size $\szbnd+1$, we build a boundary tree $\calT$ using the first $\szbnd$ elements, and use the class prediction of the last, i.e., the $\szbnd+1$st element, to take a gradient step towards minimizing the cross entropy loss. To better understand the DBT algorithm, we implement two DBT variants with one significant difference. In the first variant, DBT-v1, the gradient $\nabla_{\theta}\cc{L}$ is calculated only considering the last data point in the mini-batch. This means that even if the data points in $\calT$ are obtained via transformation through $\fntd$, they are considered constants with respect to $\theta$. This is equivalent to completely freezing the boundary tree along with the data points in nodes, and then optimizing $\fntd$. In contrast, the second variant, DBT-v2, considers points in the boundary tree as functions of $\theta$ when computing the gradient. Put differently, DBT-v2 computes $\nabla_{\theta}\cc{L}$ considering the last data point in the mini-batch, as well as the points in $\calT$. This variant can be thought of as jointly optimizing $\fntd$ along with points in the boundary tree. To the best of our understanding, DBT-v1 is what is implemented in~\cite{zoran2017learning}.

All algorithms are implemented using TensorFlow~\cite{abadi-tensorflow} and fully connected neural networks (NNets) employing ReLU activation functions. In the DBS and DBT algorithm variants, a NNet architecture of $784\to400\to400\to20$ is used to implement $\fntd$. For comparison, a neural network classifier (NNet${}^*$) with architecture $784\to400\to400\to20\to10$ is also trained. In each algorithm, we select the hyperparameter set that empirically results in the fastest convergence. Both $\szbnd$ and $\sztrn$ are set to 1000 in the DBT variants and 100 in DBS. We set $\sigma$ to 1 in DBT-v1 and to 60 in both DBT-v2 and DBS. The Adam~\cite{kingma2014adam} optimizing scheme is used with an initial learning rate of 0.0001 for DBT variants and 0.001 for DBS and NNet${}^*$. We successively decrease the learning rates by a factor of 10 after 400, 1000 and 3000 epochs. Each model is optimized until the test error no longer decreases or a maximum of 5000 epochs is reached. To compute the test errors in the DBT and DBS related experiments, we use a single BT built from all 60000 training points. Consistent with the observations made in~\cite{zoran2017learning}, we find that testing on a single boundary tree instead of a forest does not make a noticeable impact on the test error. 

\begin{figure*}%
	\centering
	\subfigure[][]{%
		\label{fig:tr_time}%
		\includegraphics[width=0.25\textwidth]{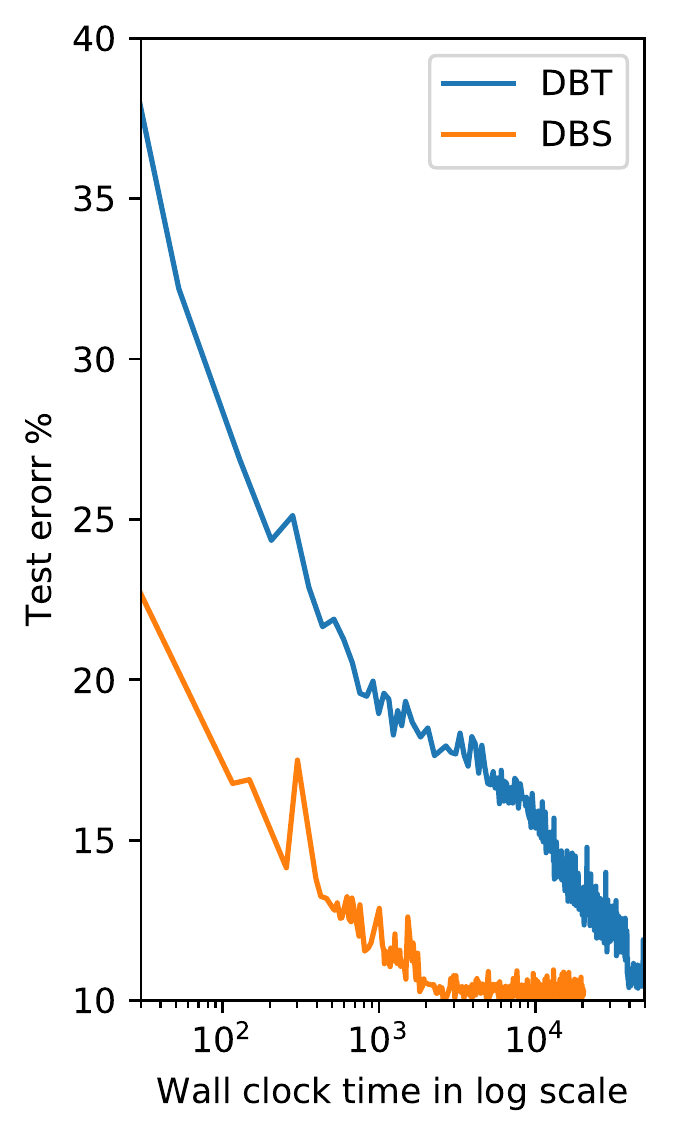}}%
	\hspace{0.05\textwidth}
	\subfigure[][]{%
		\label{fig:tsne_plot}%
		%\makebox[\textwidth][c]{\includegraphics[width=.5\textwidth]{plots/tsne.pdf}}%
		\raisebox{9mm}{\includegraphics[width=0.55\textwidth]{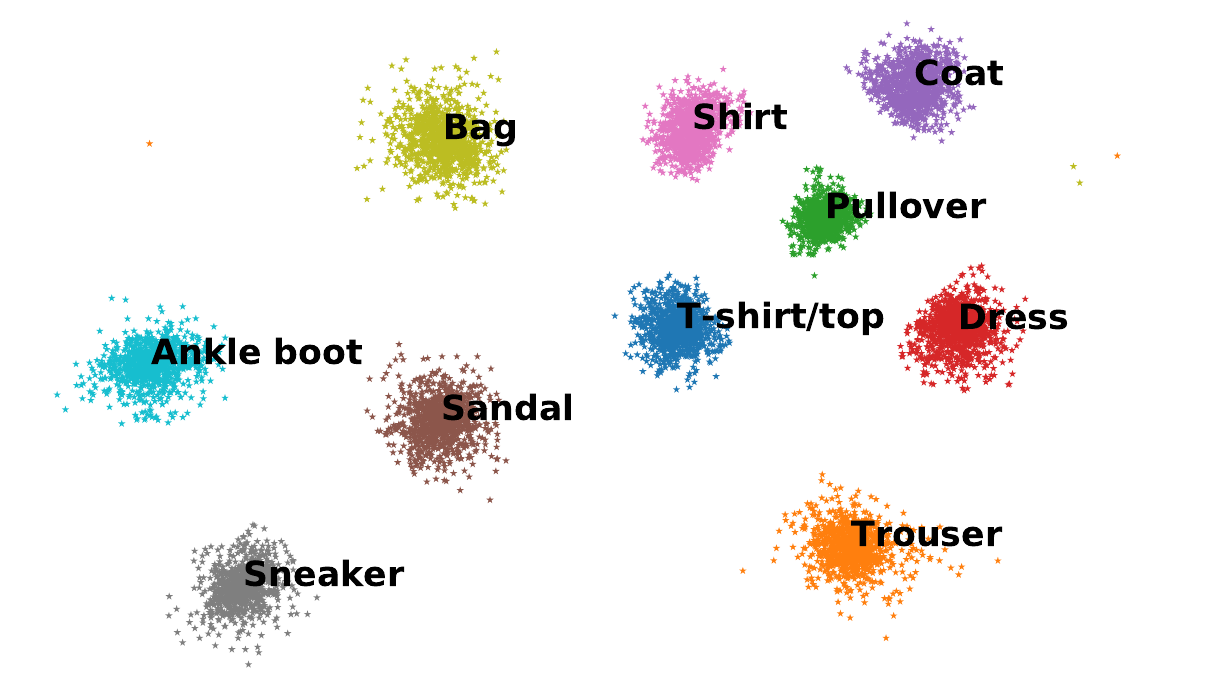}}
	}%
	\caption[]{
		Sub-figure~\subref{fig:tr_time} compares training time for DBS and DBT on the Fashion-MNIST dataset.
		Sub-figure~\subref{fig:tsne_plot} is the %t-SNE~\cite{maaten2008visualizing} 
		visualization of 2-dimensional representations learned for a random subset of Fashion-MNIST training data points. Observe that the five clusters in top-right corner are the closest to each other and are of fashion material that consist of similar prototypical examples.}
	\label{fig:time_tsne}%
\end{figure*}

The results are reported in Table~\ref{sup_err}. The `number of nodes' columns indicate the sizes of the boundary trees built using all available training data. %test errors are values presented for DBT and DBS are measured on the final BT, built using all available training data. 
Note that the test error for DBS in each dataset is on a par with the NNet${}^*$ counterpart. Also, we observe that in both the DBT and DBS results, the number of nodes in the final BT is smaller for Fashion-MNIST than for Digit-MNIST, even though the test errors are in the reverse order. We argue that this is acceptable because a higher test error does not necessarily mean a more complex class boundary. Next, one of the most beneficial aspect of DBS when compared to DBT is its computational efficiency. The authors of DBT note that, due to the discrete nature of the tree traversals for different query data points, a different TensorFlow computation graph must be built in each iteration of DBT. We note though that our implementation of the DBT algorithm mitigates this limitation by exploiting the disappearance of the traversal-dependent first term in (\ref{eq:pryc}), as described at the end of Section~\ref{sec:dbt}. However, batch-implementation of the DBT algorithm is still impracticable, which remains a significant limiting factor. The training time of two algorithms depend heavily on their implementations and the type of hardware used, but generally is higher for DBT as DBT is unable to exploit fully the benefits offered by batch-implementation dependent machine learning tools. In our simulations we observed that, to reach a given test error rate, the proposed DBS algorithm dramatically reduces the required training time. This can be observed in Figure~\ref{fig:tr_time}. Our simulations are carried out on a GeForce GTX 1060 6GB GPU and we expect the differences in the training time be even more profound if a more powerful GPU is used. 

To better illustrate the representations learned by the DBS algorithm, we set the neural network architecture to $784\to400\to400\to\mathbf{2}$, and plot the learned representations on a 2-dimensional map. As seen in Figure~\ref{fig:tsne_plot}, the transformation function learns clusters that are well separated, and are compact enough for neighbor points in same class to appeal to Euclidean distance metric. The final BT consists of 37 nodes for this version and achieves a test error of 11.5\%. (i.e. higher than in Table~\ref{sup_err} because here we use a $2$-dimensional output layer to aid in visualization)

%, which is built after training $\fntd$.% Interestingly, towards the end of the training, the size of the BS (built using $\szbnd$ number of examples) $|\calS|$ goes as low as 10, implying that only one example is needed to represent each class. This number for DBT remains at around 25. %The test errors are measured using the BS and BT built using 1000 training examples.

%most challenging, the implementation batch-wise optimization Due to the discrete nature of the path we take through the tree we need to build a different computation graph for each query node comments on training time

% \usepackage{booktabs}
% \usepackage{multirow}
\begin{table}[]
	\centering
	\caption{Test error and the number of nodes in final BT}
	\label{sup_err}

	\begin{tabular}{@{}ccccc@{}}
		\toprule
		\multirow{2}{*}{Model}              & \multicolumn{2}{c}{Digit-MNIST} & \multicolumn{2}{c}{Fashion-MNIST} \\ \cmidrule(l){2-5} 
		& Test error \%   & \# of nodes   & Test error \%    & \# of nodes    \\ \midrule
		NNet${}^*$ & 1.48            & -             & 9.8             & -              \\ \midrule
		DBT-v1                    & 2.23             & 220          & 14.2             & 505            \\ \midrule
		DBT-v2                    & 1.71             & 46          & 10.9             & 31            \\ \midrule
		DBS               & 1.52            & 29          & 10.3             &26           \\ \bottomrule
	\end{tabular}

\end{table}

%In DBS tests out of 1000 training points given to train BS, only 10 points are needed in the end. This means the neural network is doing a good job simplifying and separating classes, therefore, only one example per class is needed to represent BS. To see how much work is done by $\fntd$, we test the DBS on a shallow network with architecture $784\to400\to20$.

%The result Also, does 1 include our discussion of trying to use a shallower network for ftheta to better share the effort between the BT (at the end) and the feature extraction?

%To demonstrate the contribution of the learned function and  $\fntd$

%\subsection{Semi-supervised experiments}

%We make use of the same LN architecture used in fully connected MLP section in cite, with following exception in the last layer of E.
%We denote the activations instead  y = h(L)

%basically it  is able to reach a .89 error rate using only 1000 labeled examples in Digit-MNIST. 784-1000-500-250-250-250-10

%\input{table_ssl_err}
%<next version>

\section{Conclusion}
%%%%%%%%%%%%%%%%%%%%%%%%%%%%%%%%%%%%%%%%%%%%%%%%%%%%%%%%%%%%%%%%%%%%%%%%%%%%%%%%%
In this paper, we introduced the boundary set (BS) and differentiable boundary set (DBS) algorithms, which we build off the recently proposed boundary trees and differentiable boundary trees. The DBS procedure iteratively trains a neural network and a boundary set to efficiently learn simple representations of complex inputs. Through the proposed algorithm, we address the computational issues of differentiable boundary trees, and also improves their classification accuracy and data representability. Our algorithm is efficiently implementable on currently available software packages and is able to fully exploit the benefits offered by modern machine learning tools.

One of the original motivations of the boundary trees is the development of a learning algorithm that can quickly adapt to new training data. We note that when combined with the transform learned by the DBS algorithm, the ability to adapt quickly can prove useful in certain contexts. For example, consider a situation where the transformation function has been learned using the DBS scheme, and a BT is trained and deployed in production. Whenever new training data becomes available, the BT can easily be re-trained on the new data without adjusting the learned transformation function. This may not be possible in, for example, a neural network classifier since the network itself has to be re-trained to adapt to each new data point.

\iffalse

++ sigma squared
++ aside sigma=60 why comment
++ squared dists for distance?

\fi

%\input{resources}
%\section*{References}

\small

%\bibliographystyle{unsrt}
%\bibliography{nips_2018}

\end{document}